\pdfoutput=1
\documentclass[11pt]{article}

\usepackage[final]{acl}
\usepackage{times}
\usepackage{latexsym}
\usepackage{booktabs}

\usepackage[T1]{fontenc}
\usepackage[utf8]{inputenc}

\usepackage{microtype}

\usepackage{inconsolata}

\usepackage{graphicx}

\usepackage{algorithm}
\usepackage{algorithmic}
\usepackage{natbib}
\usepackage{multirow}
\usepackage{svg}
\usepackage{amsmath,amssymb}
\usepackage{here}
\usepackage{url}
\usepackage{comment}
\usepackage{tabularx}
\usepackage{color}
\usepackage{xspace}
\usepackage[most]{tcolorbox}

\title{Let's Put Ourselves in Sally's Shoes: Shoes-of-Others Prefilling Improves Theory of Mind in Large Language Models}

\author{Kazutoshi Shinoda ~~ Nobukatsu Hojo ~~ Kyosuke Nishida ~~ Yoshihiro Yamazaki \\
\textbf{Keita Suzuki ~~ Hiroaki Sugiyama ~~ Kuniko Saito}\\
  NTT, Inc.\\
  \texttt{kazutoshi.shinoda@ntt.com} \\}

\begin{document}
\maketitle
\begin{abstract}
Recent studies have shown that Theory of Mind (ToM) in large language models (LLMs) has not reached human-level performance yet. Since fine-tuning LLMs on ToM datasets often degrades their generalization, several inference-time methods have been proposed to enhance ToM in LLMs. However, existing inference-time methods for ToM are specialized for inferring beliefs from contexts involving changes in the world state. In this study, we present a new inference-time method for ToM, Shoes-of-Others (SoO) prefilling, which makes fewer assumptions about contexts and is applicable to broader scenarios. SoO prefilling simply specifies the beginning of LLM outputs with ``Let's put ourselves in A's shoes.'', where A denotes the target character's name. We evaluate SoO prefilling on two benchmarks that assess ToM in conversational and narrative contexts without changes in the world state and find that it consistently improves ToM across five categories of mental states. Our analysis suggests that SoO prefilling elicits faithful thoughts, thereby improving the ToM performance.
\end{abstract}

\begin{figure}[t]
    \centering
    \includegraphics[width=7.3cm]{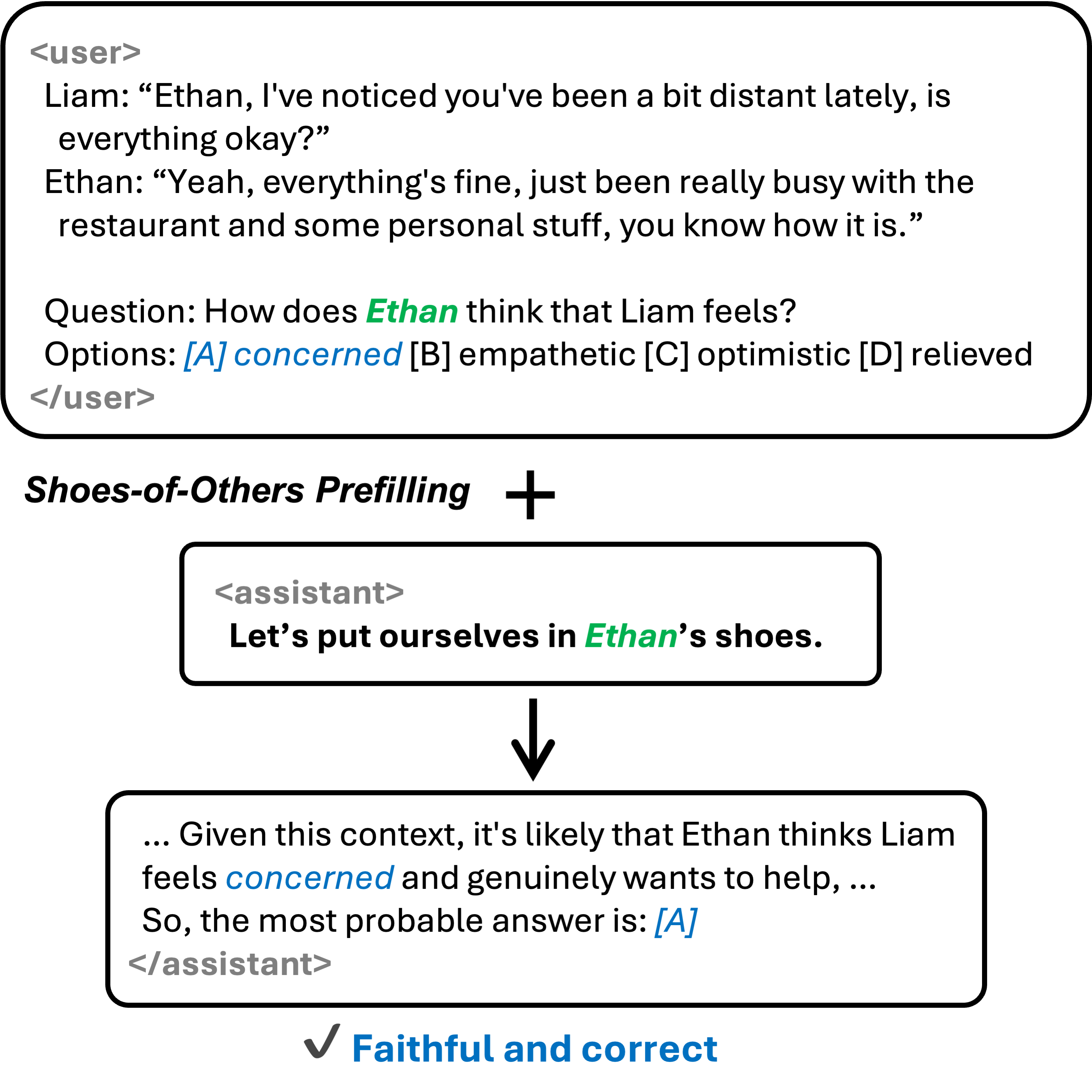}
    \caption{Shoes-of-Others prefilling specifies the beginning of outputs and then LLMs generate the continuation. The above example from ToMATO \cite{shinoda2025tomato} illustrates that Shoes-of-Others encourages the generation of faithful thoughts (i.e., the reasoning process accurately explains its prediction), thereby improving performance. See \S\ref{sec:analysis} for in-depth analyses.}
    \label{fig:overview}
\end{figure}

\section{Introduction}
Theory of Mind (ToM) is the ability to infer the mental states of others \cite{premack_woodruff_1978}.
ToM is thought to be indispensable for humans to interact with others effectively \cite{baron1985autistic}.
The ToM capability of large language models (LLMs) is also becoming necessary as LLMs are used in situations that require ToM \cite{sap-etal-2022-neural}, such as email autocomplete \cite{chen2019gmail}, empathic conversation \cite{sharma2021towards}, and persuasion \cite{wang-etal-2019-persuasion}.
Meanwhile, previous studies have shown that the ToM performance of LLMs still lags behind humans in recently proposed benchmarks \cite{sap-etal-2022-neural,ullman2023large,shapira-etal-2024-clever,kim-etal-2023-fantom,chen2024tombench,xu2024opentom,shinoda2025tomato}.
Fine-tuning LLMs on ToM datasets can improve in-distribution generalization \cite{kim-etal-2023-fantom,xu2024opentom}, but it induces overfitting and degrades out-of-distribution generalization \cite{sclar-etal-2023-minding}.

This background has led to growing interest in inference-time methods for ToM that avoid tuning model parameters and do not suffer from overfitting by design \cite{sclar-etal-2023-minding,wilf2024think,hou-etal-2024-timetom,sarangi-etal-2025-decompose}.
E.g., \citet{sclar-etal-2023-minding} designed a pipeline to track the beliefs of characters in narratives like Sally-Anne tests \cite{le-etal-2019-revisiting}.
\citet{wilf2024think} filtered contexts based on what the character in question knows before answering questions.
However, these methods are specifically designed for reasoning about \textit{beliefs} in contexts where \textit{the world state changes}\footnote{E.g., Sally puts a ball into a box and \textit{leaves} the room. Then, Anne \textit{moves} a ball into a basket. Q. Where will Sally look for the ball when she returns? \cite{baron1985autistic}}, limiting their applicability to broader scenarios.
As discussed in \citet{ma-etal-2023-towards-holistic}, various categories of mental states can be inferred through ToM in diverse scenarios.
Therefore, methods that focus only on beliefs in specific scenarios are insufficient to develop practical and human-like ToM with LLMs.

In this study, we present a new inference-time method, Shoes-of-Others (SoO) prefilling, to enhance ToM in LLMs across various mental states.
SoO prefilling specifies the beginning of LLM outputs to initiate reasoning before answering questions (\S\ref{sec:soo}) as illustrated in Figure \ref{fig:overview}.
Because SoO prefilling makes fewer assumptions about contexts, it is applicable to a broader range of settings.

We demonstrate that SoO prefilling improves ToM in LLMs for five categories of mental states (beliefs, intentions, desires, emotions, and knowledge) in conversations and narratives without changes in the world state, outperforming other methods (\S\ref{sec:exp}).
Furthermore, our analysis suggests that SoO prefilling encourages LLMs to generate more faithful thoughts than ``Let's think step-by-step'' without relying solely on lengthening thought, which may contribute to the superior performance (\S\ref{sec:analysis}).
We believe that our findings lay the foundation for developing a specialized thought process for ToM reasoning.

\section{Shoes-of-Others Prefilling}
\label{sec:soo}
SoO prefilling simply specifies the prefix of outputs with ``Let's put ourselves in \{name\}'s shoes.''
Here, the \{name\} placeholder is filled with the name of a character whose mental state is asked by a question.
We extracted the names from questions in a rule-based manner as detailed in Appendix \ref{app:name-extraction}.
Then, LLMs generate the continuation of the prefix.

Similar to \citet{wilf2024think}, we design our method inspired by perspective taking \cite{davis1996effect,ruby2004how}, which is to think from others' perspectives.
Perspective taking is thought to be necessary for successful ToM reasoning \cite{frith2006neural}.
While \citet{wilf2024think} proposed to filter out information that a character does not know from contexts based on perspective taking, this is only applicable to contexts with changes to the world state.
In contrast, SoO is designed to better guide the thought processes of LLMs with fewer assumptions about contexts.

Prefilling differs from traditional prompt-based instruction for LLMs; rather than providing instructions, it specifies the initial portion of the output and allows the model to generate the continuation.
In recent years, there has been increasing interest in research on prefilling, such as the discovery of the prefilling attack, which can induce LLMs to generate harmful outputs \cite{qi2025safety}.
In contrast, our proposed method is a prefilling strategy designed to improve ToM capabilities by prompting LLMs to adopt another’s perspective explicitly.

\section{Experiments}
\label{sec:exp}
\subsection{Experimental Setups}
\paragraph{Datasets.}
We evaluated ToM on two recently proposed benchmarks, ToMATO \cite{shinoda2025tomato} and ToMBench \cite{chen2024tombench} as these benchmarks assess ToM across broad categories of mental states in conversational and narrative contexts, respectively, without changes to the world state.
Examples are given in Appendix \ref{app:example}.
ToMATO can also evaluate the robustness to diverse personalities as reported in Appendix \ref{app:personality}.
For ToMBench, we used only the English subset.
We used accuracy as a metric because both the benchmarks are formulated as multiple-choice question answering.
Because we set the number of options for each question to four, the random baseline is 25\%.
The sizes of ToMATO and ToMBench are 5.4k and 2.4k, respectively.
The prompts used for evaluation are given in Appendix \ref{app:prompt}.

\paragraph{Methods.}
We compared the following five methods in our experiments: 1) Vanilla: zero-shot prompting. 2) (zero-shot) CoT prompting \cite{kojima2022large}: Append ``\# Answer\textbackslash n Let's think step-by-step.'' to inputs. 3) SoO prompting: Append ``Let's put ourselves in \{name\}'s shoes.'' to inputs. 4) CoT prefilling: Prefix ``Let's think step-by-step.'' to outputs. 5) SoO prefilling: Our proposed method.

\paragraph{Models.}
In our experiments, we used three open-weight LLMs (Mistral-7B-Instruct-v0.3 \cite{jiang2023mistral7b}, Llama-3-8B-Instruct, and Llama-3-70B-Instruct \cite{dubey2024llama}) and proprietary LLMs (GPT-3.5 turbo and GPT-4o mini).\footnote{Since OpenAI's models do not support a function to specify the prefix of outputs, only prompting methods were evaluated with them.}

\begin{table*}[tbp]
\small
\centering
\begin{tabular}{c|c|cccccc|cccccc}
\toprule
&& \multicolumn{6}{c|}{ToMATO} & \multicolumn{6}{c}{ToMBench} \\
\cmidrule{3-14}
Model & Method & B & I & D & E & K & Avg. & B & I & D & E & K & Avg. \\
\midrule
GPT-3.5 & Vanilla & 58.5 & 57.0 & 69.8 & 61.8 & 55.6 & 60.5 & 51.7 & 63.6 & 49.4 & 62.3 & 32.1 & 51.8\\
turbo & CoT Prompting & 72.0 & 72.9 & 77.7 & 71.5 & 74.1 & 73.7 & 56.9 & 59.3 & \textcolor{red}{44.4} & 67.9 & 36.9 & 53.1\\
 & SoO Prompting & 59.2 & \textcolor{red}{50.4} & \textcolor{red}{67.1} & 61.8 & 57.9 & \textcolor{red}{59.3} & 53.3 & \textcolor{red}{63.3} & \textcolor{red}{47.5} & 64.4 & \textcolor{red}{30.2} & \textcolor{red}{51.7}\\\midrule
GPT-4o & Vanilla & 76.2 & 79.9 & 82.4 & 76.8 & 73.3 & 77.7 & 61.7 & 72.1 & 60.6 & 71.5 & 35.1 & 60.2\\
mini & CoT Prompting & \textcolor{red}{48.1} & \textcolor{red}{46.4} & \textcolor{red}{51.7} & \textcolor{red}{62.6} & \textcolor{red}{43.8} & \textcolor{red}{50.5} & \textcolor{red}{44.9} & \textcolor{red}{30.5} & \textcolor{red}{34.4} & \textcolor{red}{46.6} & \textcolor{red}{28.4} & \textcolor{red}{36.9}\\
 & SoO Prompting & \textcolor{red}{75.7} & \textcolor{red}{79.7} & 83.7 & \textcolor{red}{75.6} & \textcolor{red}{72.3} & \textcolor{red}{77.4} & 64.4 & \textcolor{red}{60.0} & \textcolor{red}{51.9} & \textcolor{red}{71.0} & \textcolor{red}{34.3} & \textcolor{red}{56.3}\\\midrule
Mistral & Vanilla & 62.0 & 68.0 & 74.5 & 60.7 & 62.4 & 65.5 & 50.5 & 56.3 & \textbf{51.0} & 61.6 & 27.1 & 49.3\\
7B & CoT Prompting & 62.7 & \textbf{70.7} & 74.5 & 62.4 & \textbf{65.5} & 67.2 & 53.7 & 58.5 & \textcolor{red}{50.2} & \textcolor{red}{60.5} & 30.6 & 50.7\\
 & SoO Prompting & 63.9 & 68.9 & \textbf{76.5} & 62.7 & 63.3 & 67.1 & 53.3 & 57.9 & \textcolor{red}{50.4} & 61.6 & 29.1 & 50.5\\
 & CoT Prefilling & \textcolor{red}{61.5} & \textcolor{red}{67.6} & \textcolor{red}{72.7} & 61.8 & \textcolor{red}{62.2} & \textcolor{red}{65.1} & 53.0 & \textcolor{red}{55.3} & \textcolor{red}{41.5} & \textcolor{red}{59.9} & 30.8 & \textcolor{red}{48.1}\\
 & SoO Prefilling & \textbf{64.9} & 70.0 & 75.6 & \textbf{63.0} & 64.5 & \textbf{67.6} & \textbf{56.2} & \textbf{58.9} & \textcolor{red}{47.3} & \textbf{63.7} & \textbf{34.5} & \textbf{52.1}\\\midrule
Llama3 & Vanilla & 54.2 & 56.1 & 60.2 & 57.0 & 47.1 & 54.9 & 48.7 & 56.0 & 49.2 & 61.2 & 31.1 & 49.2\\
8B & CoT Prompting & \textcolor{red}{26.0} & \textcolor{red}{26.2} & \textcolor{red}{22.0} & \textcolor{red}{28.9} & \textcolor{red}{24.7} & \textcolor{red}{25.6} & \textcolor{red}{46.1} & \textcolor{red}{41.9} & \textcolor{red}{36.9} & \textcolor{red}{51.8} & \textcolor{red}{28.0} & \textcolor{red}{40.9}\\
 & SoO Prompting & \textcolor{red}{51.6} & 57.9 & \textcolor{red}{51.1} & \textcolor{red}{55.6} & \textcolor{red}{41.6} & \textcolor{red}{51.6} & \textcolor{red}{47.9} & \textcolor{red}{47.1} & \textcolor{red}{45.0} & \textcolor{red}{57.6} & 32.2 & 46.0\\
 & CoT Prefilling & 64.1 & 65.3 & 71.0 & 60.8 & 58.9 & 64.0 & 55.3 & 65.2 & 51.9 & 63.4 & 37.2 & 54.6\\
 & SoO Prefilling & \textbf{67.2} & \textbf{69.2} & \textbf{73.4} & \textbf{65.7} & \textbf{62.0} & \textbf{67.5} & \textbf{61.1} & \textbf{70.8} & \textbf{59.0} & \textbf{66.6} & \textbf{38.3} & \textbf{59.2}\\\midrule
Llama3 & Vanilla & 81.7 & 85.3 & 85.9 & 80.5 & 73.5 & 81.4 & 73.6 & 79.8 & 58.5 & 71.9 & 45.9 & 66.0\\
70B & CoT Prompting & \textcolor{red}{80.5} & \textcolor{red}{85.2} & 86.7 & 81.3 & 74.1 & 81.6 & \textcolor{red}{68.2} & \textcolor{red}{78.8} & \textcolor{red}{54.0} & \textcolor{red}{69.6} & 50.6 & \textcolor{red}{64.2}\\
 & SoO Prompting & 81.9 & 86.2 & \textbf{87.6} & 82.2 & 75.6 & 82.7 & \textcolor{red}{73.3} & 80.7 & \textbf{59.4} & 72.8 & 51.9 & 67.6\\
 & CoT Prefilling & \textcolor{red}{79.9} & \textcolor{red}{83.9} & \textcolor{red}{84.2} & \textcolor{red}{78.6} & \textcolor{red}{73.4} & \textcolor{red}{80.0} & \textcolor{red}{71.1} & \textcolor{red}{69.1} & \textcolor{red}{51.5} & \textcolor{red}{65.8} & \textbf{49.0} & \textcolor{red}{61.3}\\
 & SoO Prefilling & \textbf{82.2} & \textbf{86.9} & 87.4 & \textbf{82.4} & \textbf{76.7} & \textbf{83.1} & \textbf{80.5} & \textbf{80.8} & 57.9 & \textbf{73.0} & 47.9 & \textbf{68.0}\\
\bottomrule
\end{tabular}
\caption{First-order ToM performance on ToMATO and ToMBench (\%). Accuracies averaged over three runs are reported. B: belief, I: intention, D: desire, E: emotion, K: knowledge. The best scores among the five inference-time methods for each model are \textbf{boldfaced}.
Degraded scores compared to Vanilla are in \textcolor{red}{red}.
}
\label{tab:results-first-tom}
\end{table*}

\begin{table*}[tbp]
\small
\centering
\begin{tabular}{c|c|cccccc|cccccc}
\toprule
& & \multicolumn{6}{c|}{True Belief} & \multicolumn{6}{c}{False Belief} \\
\cmidrule{3-14}
Model & Method & B & I & D & E & K & Avg. & B & I & D & E & K & Avg. \\
\midrule
GPT-3.5 & Vanilla & 54.9 & 54.1 & 57.6 & 55.4 & 55.6 & 55.5 & 42.5 & 35.5 & 47.0 & 37.5 & 45.7 & 41.7\\
turbo & CoT Prompting & 66.4 & 66.9 & 70.3 & 69.0 & 68.8 & 68.3 & 57.0 & 46.7 & 62.0 & 57.5 & 60.5 & 56.7\\
 & SoO Prompting & 56.4 & \textcolor{red}{52.4} & 58.5 & 57.0 & 56.6 & 56.2 & \textcolor{red}{40.1} & 38.5 & \textcolor{red}{44.9} & 40.2 & 47.5 & 42.2\\\midrule
GPT-4o & Vanilla & 69.1 & 70.7 & 77.2 & 71.7 & 73.7 & 72.5 & 60.1 & 47.8 & 71.9 & 71.7 & 58.6 & 62.0\\
mini & CoT Prompting & \textcolor{red}{35.2} & \textcolor{red}{46.3} & \textcolor{red}{45.4} & \textcolor{red}{51.9} & \textcolor{red}{36.5} & \textcolor{red}{43.0} & \textcolor{red}{32.9} & \textcolor{red}{35.2} & \textcolor{red}{47.5} & \textcolor{red}{52.0} & \textcolor{red}{31.5} & \textcolor{red}{39.8}\\
 & SoO Prompting & 69.2 & \textcolor{red}{70.5} & \textcolor{red}{71.7} & 74.9 & \textcolor{red}{72.9} & \textcolor{red}{71.8} & 62.0 & \textcolor{red}{47.5} & \textcolor{red}{66.5} & \textcolor{red}{70.9} & \textcolor{red}{56.8} & \textcolor{red}{60.7}\\\midrule
Mistral & Vanilla & 57.3 & 61.9 & 61.4 & 61.6 & 62.4 & 60.9 & 42.9 & 39.9 & 48.9 & 48.6 & 50.8 & 46.2\\
7B & CoT Prompting & \textcolor{red}{55.2} & \textbf{64.0} & \textbf{64.2} & 63.1 & \textcolor{red}{61.9} & \textbf{61.7} & \textcolor{red}{41.9} & 41.5 & \textbf{57.0} & 48.8 & 53.1 & 48.5\\
 & SoO Prompting & \textbf{58.9} & 62.5 & 62.1 & \textcolor{red}{60.8} & \textbf{64.3} & \textbf{61.7} & 44.2 & 42.9 & 53.2 & \textcolor{red}{47.0} & \textcolor{red}{49.4} & 47.3\\
 & CoT Prefilling & \textcolor{red}{54.8} & 62.9 & 62.5 & 63.5 & \textcolor{red}{62.1} & 61.2 & 44.4 & \textbf{43.4} & 55.7 & 50.9 & 52.3 & 49.4\\
 & SoO Prefilling & \textcolor{red}{52.0} & 62.4 & 61.7 & \textbf{64.1} & 63.8 & \textcolor{red}{60.8} & \textbf{47.5} & 43.2 & 50.4 & \textbf{54.6} & \textbf{56.6} & \textbf{50.5}\\\midrule
Llama3 & Vanilla & 39.8 & 45.6 & 46.3 & 46.9 & 40.4 & 43.8 & 34.5 & 29.5 & 35.4 & 37.0 & 27.8 & 32.8\\
8B & CoT Prompting & \textcolor{red}{25.6} & \textcolor{red}{23.9} & \textcolor{red}{25.0} & \textcolor{red}{27.3} & \textcolor{red}{23.4} & \textcolor{red}{25.1} & \textcolor{red}{25.6} & \textcolor{red}{21.6} & \textcolor{red}{29.3} & \textcolor{red}{27.0} & \textcolor{red}{22.8} & \textcolor{red}{25.3}\\
 & SoO Prompting & \textcolor{red}{36.1} & 47.2 & \textcolor{red}{44.9} & 47.5 & \textcolor{red}{35.0} & \textcolor{red}{42.1} & \textcolor{red}{32.4} & 30.6 & 39.7 & 41.2 & \textcolor{red}{27.6} & 34.3\\
 & CoT Prefilling & 58.6 & 60.6 & 61.2 & 58.0 & 60.8 & 59.8 & \textbf{49.5} & \textbf{46.2} & 53.2 & 54.3 & \textbf{52.9} & \textbf{51.2}\\
 & SoO Prefilling & \textbf{60.1} & \textbf{61.4} & \textbf{64.4} & \textbf{59.9} & \textbf{61.8} & \textbf{61.5} & 48.1 & 43.4 & \textbf{55.1} & \textbf{56.4} & 50.2 & 50.6\\\midrule
Llama3 & Vanilla & 73.7 & 76.1 & 78.7 & 75.8 & 70.2 & 74.9 & 60.6 & 57.1 & 67.3 & 70.1 & 58.4 & 62.7\\
70B & CoT Prompting & 77.0 & 76.5 & 78.9 & 79.7 & 75.7 & 77.6 & \textbf{63.7} & \textbf{62.3} & 68.3 & 74.8 & 63.2 & 66.5\\
 & SoO Prompting & 76.4 & \textbf{79.4} & 81.1 & 77.9 & \textbf{77.2} & 78.4 & 62.7 & 61.2 & 69.2 & 72.2 & \textbf{64.2} & 65.9\\
 & CoT Prefilling & 75.7 & \textcolor{red}{76.0} & 79.1 & 78.4 & 72.1 & 76.3 & \textcolor{red}{60.2} & 58.2 & \textbf{71.7} & 77.2 & 62.1 & 65.9\\
 & SoO Prefilling & \textbf{79.5} & 77.8 & \textbf{81.2} & \textbf{81.5} & 76.3 & \textbf{79.3} & 62.6 & 61.2 & 70.9 & \textbf{78.0} & 62.1 & \textbf{67.0}\\
\bottomrule
\end{tabular}
\caption{Second-order ToM performance on ToMATO (\%). Accuracies averaged over three runs are reported.
}
\label{tab:results-second-tom}
\end{table*}

\begin{table*}[tbp]
\centering
\small
\begin{tabular}{c|c|c|c}
\toprule
Method & Prefix & ToMATO & ToMBench \\
\midrule
SoO Prefilling & Let's put ourselves in \{name\}'s shoes. & 62.8 & 63.2\\
- name & Let's put ourselves in others' shoes. & 29.4 & 43.0\\
- name & Let's put ourselves in shoes of others. & 16.2 & 28.8\\
\bottomrule
\end{tabular}
\caption{Ablation study with Llama-3-8B-Instruct.}
\label{tab:ablation}
\end{table*}

\subsection{Results}
\paragraph{First-order Theory of Mind.}
First-order ToM refers to reasoning about first-order mental states, e.g., ``A thinks/will/wants/feels/knows X'' corresponds to first-order belief/intention/desire/emotion/knowledge, respectively.

As shown in Table \ref{tab:results-first-tom}, SoO prefilling was consistently effective across the five mental states for both conversation (ToMATO) and narrative (ToMBench) inputs with some exceptions.
On the other hand, the prompting methods were not always effective and tended to lower scores compared to Vanilla.

\paragraph{Second-order Theory of Mind.}
Second-order ToM refers to reasoning about second-order mental states, e.g., ``A thinks that B thinks/will/wants/feels/knows Y'' corresponds to second-order belief about belief/intention/desire/emotion/knowledge, respectively.
False beliefs are beliefs that differ from actual beliefs. E.g., ``A thinks that B feels Y, while B feels X'' corresponds to true (false) beliefs about emotion if X $=$ ($\neq$) Y.

As shown in Table \ref{tab:results-second-tom}, SoO prefilling outperformed other methods for Llama3 70B on average (Avg.).
In addition, while prompting methods often degraded the scores compared to Vanilla for proprietary and open-weight models, SoO prefilling consistently improved the scores.
Notably, SoO prefilling was more effective on false belief tasks than on true belief subsets.

\def\Height{3.2cm}
\begin{figure}[t]
    \centering
    \begin{tabular}{c}
    \includegraphics[height=\Height]{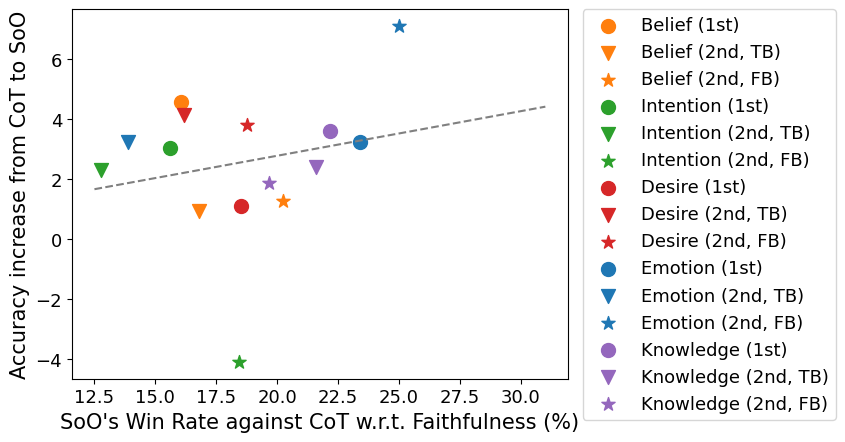} \\
    (a) ToMATO\\
    \includegraphics[height=\Height]{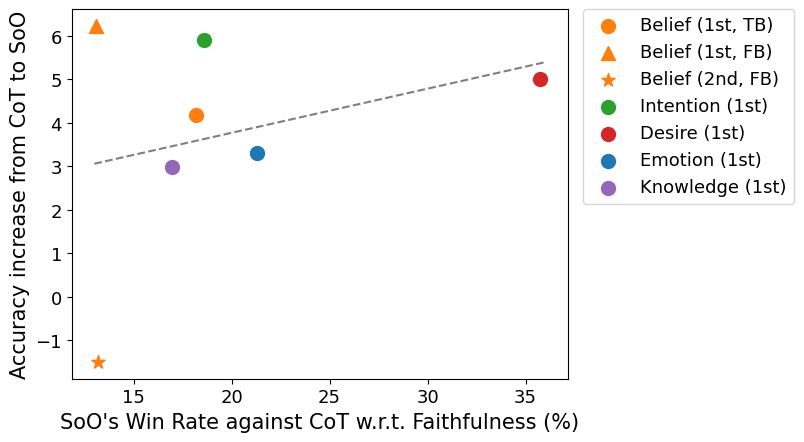} \\
    (b) ToMBench\\
    \end{tabular}
    \caption{Correlation analysis of accuracy and faithfulness for Llama-3-8B-Instruct. The correlation between the two is positive on both benchmarks.}
    \label{fig:acc-winrate}
\end{figure}

\begin{figure}[t]
    \centering
    \begin{tabular}{c}
    \includegraphics[height=\Height]{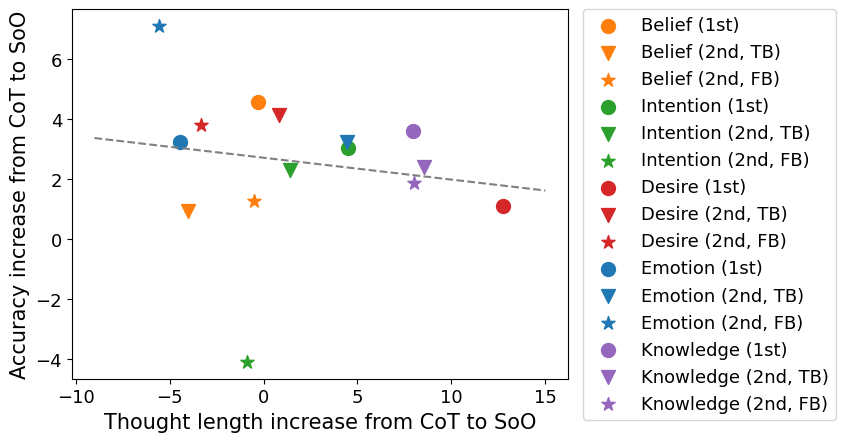} \\
    (a) ToMATO\\
    \includegraphics[height=\Height]{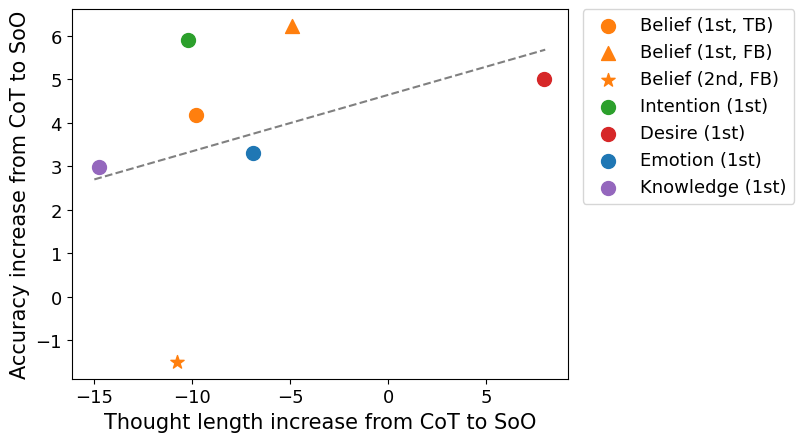} \\
    (b) ToMBench\\
    \end{tabular}
    \caption{Correlation analysis of accuracy and thought length for Llama-3-8B-Instruct. The correlation between the two is not necessarily positive.}
    \label{fig:acc-length}
\end{figure}

\paragraph{Ablation Study.}
To see the effect of including the name of a character in SoO prefilling, we conducted an ablation study.
As shown in Table \ref{tab:ablation}, including the names is consistently effective for the two benchmarks.
Explicitly specifying the names in prefixes would be necessary for properly guiding the thought processes for ToM reasoning.

\section{Analysis}
\label{sec:analysis}
\paragraph{Why does SoO prefilling outperform CoT prefilling?}
Chain-of-Thought often suffers from unfaithfulness, i.e., thoughts do not accurately explain final predictions of LLMs \cite{lyu-etal-2023-faithful,lanham2023measuring}.
An example of unfaithful thoughts is shown in Figure~\ref{tab:example-tomato}.
We hypothesize that SoO prefilling outperformed other methods in \S\ref{sec:exp} by mitigating the unfaithfulness.

To verify this hypothesis, we first utilized LLM-as-a-judge \cite{zheng2023judging} with GPT-4o mini to compare the faithfulness of SoO and CoT prefilling.
Namely, we extended a prompt for pairwise comparison \cite{zheng2023judging} to compare the faithfulness of thoughts as in Appendix \ref{app:prompt-judge}.
For LLM-as-a-judge, we used examples where Llama3 8B made correct predictions with both CoT and SoO prefilling for fair comparison.
Then, we conducted correlation analyses for accuracy increase and win rate w.r.t. faithfulness.

As shown in Figure \ref{fig:acc-winrate}, the win rates are positively correlated with the accuracy increases on the two benchmarks.
This trend is also observed in other models as reported in Appendix \ref{app:acc-winrate-length}.
These findings suggest that SoO prefilling mitigates the unfaithfulness of thoughts, thereby improving ToM performance.
Such analyses of the relationship between intervention in LLM output, faithfulness of thought, and ToM are a relatively unexplored area in LLM ToM research.

\paragraph{Does SoO prefilling Scale Test-Time Compute to Improve ToM?}
To answer this question, we also conducted correlation analyses of accuracy and thought length increase from CoT to SoO prefilling.
As shown in Figure \ref{fig:acc-length}, we find that the thought length increase does not necessarily correlate positively with the accuracy increase.
This result implies that SoO prefilling does not rely solely on making thoughts longer, i.e., scaling test-time compute \cite{snell2024scaling}, to improve ToM.

\section{Discussion \& Conclusion}
\paragraph{Summary.}
We proposed SoO prefilling to improve ToM in LLMs without sacrificing its applicability.
Our experiments on two benchmarks showed that our method improved first- and second-order ToM across five mental states in most cases.
Our analysis implied that our method mitigated the unfaithfulness of thoughts, thereby improving ToM.
Identifying other crucial factors for ToM is future work.

\paragraph{Relevance to ASD Research.}
This study found that the ToM performance of LLMs can be enhanced by explicitly prompting them to engage in perspective-taking.
This finding is consistent with previous research on autism spectrum disorder (ASD), which has shown that while adolescents with ASD tend to underperform in social cognition tasks compared to typically developing individuals when not prompted, their performance becomes comparable when perspective-taking is explicitly encouraged \cite{callenmark2014explicit}.
Moreover, the finding that prompting LLMs to engage in explicit reasoning before answering (explicit ToM) leads to improved performance compared to vanilla prompting (implicit ToM) is also consistent with findings from ASD research \cite{schuwerk2015implicit}.
Advancing our understanding of ToM through such comparisons between humans and LLMs remains an important direction for future research.

\section{Limitations}
One of the limitations of our work is that our proposed method is not evaluated with proprietary LLMs.
Because our SoO prefilling needs to specify the prefix of LLM outputs, it can not be applied to OpenAI models such as GPT-4 \cite{openai2023gpt}.
If OpenAI supports a function to specify the prefix of outputs and to generate the continuation, our SoO prefilling can be used with them.
Second, the performance gain drawn by SoO prefilling depends on models and data.
This may be due to suboptimal designs of prompt formats.
Third, LLM-as-a-judge we used for the analysis may suffer from biases, such as positions and verbosity.
This may undermine the reliability of the insights drawn from our analysis.
To alleviate this concern, we randomly selected the positions of two AI assistants when using LLM-as-a-judge.
In addition, we showed that the thought lengths are not positively correlated with the win rates as in Appendix \ref{app:winrate-length}.

\section{Ethical Considerations}

AI's ToM could predict human beliefs, intentions, and emotions.
First, This raises concerns about unethical use cases (e.g., influencing decisions in social media, marketing, or political campaigns).
Malicious actors may misuse AI's ToM to deceive or persuade humans in such use cases.
Second, AI's ToM could involve collecting and analyzing sensitive data.
It could lead to intrusive surveillance or unwanted profiling (e.g., predicting mental states in schools or workplaces without consent).
Third, if AI systems misinterpret human mental states and make unfavorable decisions based on them, humans may suffer a disadvantage, especially in critical applications (e.g., healthcare, and education).
If humans over-rely on AI's ToM, it may reinforce biases and lead to false positives.
We would need to be careful when using AI's ToM not to cause harm to humans.

\bibliography{custom}

\appendix

\section{Prompt}
\label{app:prompt}
\subsection{Prompts for Evaluation}
\label{app:prompt-eval}
\begin{tcolorbox}[title=Chat Messages for Evaluation on ToMATO,
colback=white,
colframe=black,
colbacktitle=white,
coltitle=black,
standard jigsaw,
opacityback=0,
breakable,
fonttitle=\bfseries]
messages = [\\
\{``role": ``system", ``content": ``You are an expert at understanding human communication. Please leverage the information provided and choose the most probable answer to the question from the options. Output your final answer by strictly following this format: [A], [B], [C], or [D]"\},\\
\{``role": ``user", ``content": ``````\# Transcript\\
\{\{conversation\}\}\\\\
\# Question\\
\{\{question\}\}\\
\\
\# Options\\
{[A]} \{\{option1\}\}\\
{[B]} \{\{option2\}\}\\
{[C]} \{\{option3\}\}\\
{[D]} \{\{option4\}\}''''''\}\\
]
\end{tcolorbox}

\begin{tcolorbox}[title=Chat Messages for Evaluation on ToMBench,
colback=white,
colframe=black,
colbacktitle=white,
coltitle=black,
standard jigsaw,
opacityback=0,
breakable,
fonttitle=\bfseries]
messages = [\\
\{``role": ``system", ``content": ``You are an expert at understanding human communication. Please leverage the information provided and choose the most probable answer to the question from the options. Output your final answer by strictly following this format: [A], [B], [C], or [D]"\},\\
\{``role": ``user", ``content": ``````\# Context\\
\{\{context\}\}\\\\
\# Question\\
\{\{question\}\}\\
\\
\# Options\\
{[A]} \{\{option1\}\}\\
{[B]} \{\{option2\}\}\\
{[C]} \{\{option3\}\}\\
{[D]} \{\{option4\}\}''''''\}\\
]
\end{tcolorbox}

\subsection{Prompts for LLM-as-a-judge}
\label{app:prompt-judge}

We randomly selected the positions of two AI assistants to mitigate the position bias of LLM-as-a-judge \cite{zheng2023judging}.

\begin{tcolorbox}[title=Chat Messages for LLM-as-a-judge to Measure Faithfulness,
colback=white,
colframe=black,
colbacktitle=white,
coltitle=black,
standard jigsaw,
opacityback=0,
breakable,
fonttitle=\bfseries]
messages = [\\
        \{``role'': ``system'', ``content'': ``````Please act as an impartial judge and evaluate the quality of the responses provided by two AI assistants to the context, user question, and options displayed below. Your evaluation should consider the truthfulness of the responses. You will be given assistant X’s answer, and assistant Y’s answer. Your job is to evaluate which assistant’s answer is more faithful. You should independently solve the user question step-by-step first. Then compare both assistants’ answers with your answer. Identify and correct any mistakes. Avoid any position biases and ensure that the order in which the responses were presented does not influence your decision. Do not allow the length of the responses to influence your evaluation. Do not favor certain names of the assistants. Be as objective as possible. After providing your explanation, output your final verdict by strictly following this format: "[[X]]" if assistant X is better, "[[Y]]" if assistant Y is better, and "[[Z]]" for a tie.''''''\},
        \{'role': 'user', 'content': ``````[Context]\\
\{\{context\}\}\\
\\
{[User Question]}\\
\{\{question\}\}

[Options]
{[A]} \{\{option1\}\}\\
{[B]} \{\{option2\}\}\\
{[C]} \{\{option3\}\}\\
{[D]} \{\{option4\}\}\\

{[The Start of Assistant X’s Answer]}\\
\{\{answer\_a\}\}\\
{[The End of Assistant X’s Answer]}\\
\\
{[The Start of Assistant Y’s Answer]}\\
\{\{answer\_b\}\}\\
{[The End of Assistant Y’s Answer]}\\
'''''']
\end{tcolorbox}

\section{Experiments}
\subsection{Experimental Setups}
\label{app:exp-setup}
Hyperparameters used in our experiments are listed in Table \ref{tab:hypara}.

\begin{table}[h]
    \centering
    \begin{tabular}{c|c}
    \toprule
    Hyperparameter & Value \\
    \midrule
    do\_sample & True \\
    top\_p & 0.9 \\
    temperature & 0.6 \\
    max\_new\_tokens & 1024\\
    \bottomrule
    \end{tabular}
    \caption{Hyperparameters}
    \label{tab:hypara}
\end{table}

For open-weight models, the three LLMs used in our experiments are as follows: Mistral-7B-Instruct-v0.3\footnote{\url{https://huggingface.co/mistralai/Mistral-7B-Instruct-v0.3}}, Llama-3-8B-Instruct\footnote{\url{https://huggingface.co/meta-llama/Meta-Llama-3-8B-Instruct}}, and Llama-3-70B-Instruct\footnote{\url{https://huggingface.co/meta-llama/Meta-Llama-3-70B-Instruct}} \cite{dubey2024llama}.
We used 4-bit quantization for the three to reduce computational costs.
We quantized these LLMs with bitsandbytes\footnote{\url{https://github.com/bitsandbytes-foundation/bitsandbytes}}.
For proprietary models, we used gpt-3.5-turbo-0125 and gpt-4o-mini-2024-07-18.

For datasets, while ToMBench has questions with two options, we filter them out for simplicity.
We used only the English subset of ToMBench.

\subsection{Name Extraction}
\label{app:name-extraction}
Our SoO prefilling uses the name of a character whose mental state is asked by a question.
In ToMATO, the names of characters can be deterministically determined because the questions were generated in a rule-based manner.
For ToMBench, the questions were written manually, so we extracted the names from the questions with a rule-based method.
We used part of speech tagging and dependency parsing with the ``en\_core\_web\_sm'' pipeline in spaCy\footnote{https://spacy.io/} to identify the names of target characters.
We excluded questions where this approach failed.
While we employed the rule-based approach for name extraction, LLM-based approaches, such as in-context learning, can also be useful.
Investigating the performance of LLM-based name extraction is future work.

\subsection{Robustness to Personality Traits}
\label{app:personality}
We also evaluated the robustness of ToM in LLMs to personality traits.
The results for first- and second-order ToM are given in Tables \ref{tab:results-first-personality} and \ref{tab:results-second-personality}, respectively.
For both cases, SoO prefilling consistently improved the ToM performance of LLMs for each factor of the big five personality traits of characters, without overfitting to some specific personalities.

\begin{table*}
\small
\centering
\begin{tabular}{c|c|cc|cc|cc|cc|cc}
\toprule
& & \multicolumn{2}{c|}{O} & \multicolumn{2}{c|}{C} & \multicolumn{2}{c|}{E} & \multicolumn{2}{c|}{A} & \multicolumn{2}{c}{N}\\
Model & Method & high & low & high & low & high & low & high & low & high & low \\
\midrule
GPT-3.5 & Vanilla & 60.4 & 59.6 & 61.5 & 57.2 & 60.4 & 59.8 & 60.7 & 59.6 & 56.3 & 60.9\\
turbo & CoT Prompting & 72.2 & 75.3 & 75.8 & 69.2 & 76.6 & 70.7 & 76.3 & 71.3 & 70.4 & 74.2\\
 & SoO Prompting & 58.7 & 58.9 & 59.8 & 56.8 & 59.3 & 58.3 & 60.4 & 57.5 & 56.8 & 59.2\\\midrule
GPT-4o & Vanilla & 77.1 & 78.2 & 78.7 & 75.4 & 78.7 & 76.5 & 79.0 & 76.4 & 75.3 & 78.0\\
mini & CoT Prompting & 48.5 & 52.2 & 50.6 & 48.9 & 51.4 & 48.8 & 49.7 & 50.4 & 48.7 & 50.4\\
 & SoO Prompting & 76.4 & 78.1 & 78.0 & 75.5 & 78.3 & 76.1 & 78.5 & 76.1 & 75.3 & 77.6\\\midrule
Mistral & Vanilla & 65.3 & 65.3 & 66.2 & 63.5 & 66.1 & 64.5 & 67.0 & 63.9 & 61.8 & 66.0\\
7B & CoT Prompting & 67.3 & 66.7 & 67.4 & 66.4 & 67.8 & 66.3 & 68.8 & 65.7 & 65.0 & 67.5\\
 & SoO Prompting & 66.5 & 67.2 & 67.7 & 64.9 & 68.1 & 65.6 & 69.3 & 64.8 & 64.0 & 67.4\\
 & CoT Prefilling & 64.7 & 65.3 & 66.3 & 62.3 & 65.8 & 64.1 & 66.8 & 63.5 & 60.6 & 65.9\\
 & SoO Prefilling & 67.6 & 67.1 & 67.8 & 66.6 & 68.1 & 66.8 & 68.6 & 66.4 & 63.8 & 68.1\\\midrule
Llama3 & Vanilla & 55.0 & 54.0 & 56.8 & 50.3 & 54.5 & 54.7 & 55.4 & 54.0 & 48.3 & 55.9\\
8B & CoT Prompting & 25.3 & 26.0 & 26.0 & 24.9 & 25.2 & 26.0 & 24.9 & 26.2 & 24.4 & 25.9\\
 & SoO Prompting & 51.1 & 52.0 & 53.0 & 48.4 & 52.2 & 50.8 & 52.1 & 51.0 & 48.1 & 52.2\\
 & CoT Prefilling & 62.7 & 65.2 & 65.3 & 60.7 & 63.6 & 63.9 & 65.1 & 62.7 & 57.8 & 65.1\\
 & SoO Prefilling & 67.5 & 66.9 & 68.2 & 65.5 & 68.0 & 66.6 & 68.8 & 66.0 & 63.4 & 68.1\\\midrule
Llama3 & Vanilla & 81.2 & 81.2 & 82.4 & 78.8 & 82.5 & 80.0 & 83.5 & 79.4 & 78.7 & 81.7\\
70B & CoT Prompting & 81.5 & 81.1 & 82.4 & 79.3 & 82.9 & 79.9 & 83.5 & 79.6 & 78.7 & 81.9\\
 & SoO Prompting & 82.8 & 82.1 & 83.3 & 80.9 & 84.3 & 80.8 & 84.9 & 80.6 & 80.8 & 82.9\\
 & CoT Prefilling & 79.8 & 80.0 & 81.1 & 77.3 & 80.6 & 79.2 & 81.4 & 78.6 & 76.6 & 80.6\\
 & SoO Prefilling & 83.3 & 82.6 & 83.9 & 81.2 & 83.5 & 82.5 & 84.4 & 81.9 & 80.7 & 83.5\\
\bottomrule
\end{tabular}
\caption{First-order ToM Performance (\%) for each factor of big five personality traits of characters on ToMATO. For each factor (O=openness to experience, C=conscientiousness, E=extraversion, A=agreeableness, N=neuroticism), the scores on two subsets (the corresponding factor is high and low) averaged over three runs are reported.
}
\label{tab:results-first-personality}
\end{table*}

\begin{table*}
\small
\centering
\begin{tabular}{c|c|cc|cc|cc|cc|cc}
\toprule
& & \multicolumn{2}{c|}{O} & \multicolumn{2}{c|}{C} & \multicolumn{2}{c|}{E} & \multicolumn{2}{c|}{A} & \multicolumn{2}{c}{N}\\
Model & Method & high & low & high & low & high & low & high & low & high & low \\
\midrule
GPT-3.5 & Vanilla & 52.4 & 50.2 & 52.5 & 50.1 & 51.8 & 51.2 & 53.1 & 50.2 & 49.0 & 52.2\\
turbo & CoT Prompting & 66.3 & 62.7 & 66.5 & 62.6 & 65.0 & 65.2 & 66.5 & 63.7 & 59.2 & 66.4\\
 & SoO Prompting & 52.2 & 51.8 & 52.8 & 50.8 & 52.6 & 51.1 & 53.2 & 51.0 & 49.2 & 52.7\\\midrule
GPT-4o & Vanilla & 69.8 & 69.1 & 71.7 & 65.7 & 69.8 & 69.1 & 70.1 & 69.0 & 62.8 & 71.0\\
mini & CoT Prompting & 42.7 & 40.4 & 42.4 & 41.0 & 41.6 & 42.5 & 42.2 & 41.6 & 39.8 & 42.4\\
 & SoO Prompting & 69.5 & 67.2 & 70.2 & 66.0 & 69.0 & 68.2 & 70.1 & 67.4 & 63.4 & 69.9\\\midrule
Mistral & Vanilla & 57.8 & 54.6 & 57.5 & 55.3 & 57.2 & 55.7 & 59.1 & 54.4 & 56.3 & 56.7\\
7B & CoT Prompting & 58.8 & 56.1 & 58.8 & 56.2 & 58.2 & 57.2 & 60.4 & 55.4 & 57.0 & 58.1\\
 & SoO Prompting & 58.1 & 56.4 & 58.3 & 56.3 & 57.6 & 57.5 & 59.5 & 55.7 & 57.3 & 57.6\\
 & CoT Prefilling & 58.8 & 55.8 & 59.0 & 55.7 & 58.1 & 57.2 & 60.3 & 55.3 & 55.9 & 58.2\\
 & SoO Prefilling & 58.7 & 56.5 & 59.0 & 56.2 & 58.4 & 57.2 & 60.5 & 55.5 & 57.3 & 58.1\\\midrule
Llama3 & Vanilla & 41.3 & 39.4 & 42.4 & 37.6 & 40.0 & 41.9 & 42.7 & 38.7 & 37.1 & 41.4\\
8B & CoT Prompting & 25.4 & 24.6 & 25.4 & 24.6 & 24.8 & 25.6 & 24.2 & 26.0 & 22.8 & 25.6\\
 & SoO Prompting & 39.8 & 39.9 & 40.7 & 38.3 & 38.9 & 41.5 & 40.2 & 39.5 & 37.4 & 40.3\\
 & CoT Prefilling & 57.4 & 57.1 & 59.0 & 54.5 & 57.9 & 56.3 & 57.9 & 56.8 & 53.5 & 58.2\\
 & SoO Prefilling & 58.6 & 57.8 & 59.9 & 55.5 & 59.0 & 57.2 & 58.9 & 57.8 & 54.4 & 59.2\\\midrule
Llama3 & Vanilla & 71.5 & 70.7 & 73.4 & 67.4 & 72.3 & 69.3 & 72.9 & 69.6 & 66.1 & 72.4\\
70B & CoT Prompting & 74.9 & 72.8 & 75.8 & 71.3 & 74.5 & 73.5 & 75.2 & 73.1 & 68.7 & 75.4\\
 & SoO Prompting & 75.1 & 73.9 & 76.6 & 71.2 & 75.2 & 73.7 & 75.6 & 73.7 & 69.7 & 75.8\\
 & CoT Prefilling & 74.4 & 70.2 & 75.0 & 69.4 & 73.3 & 72.4 & 74.9 & 71.1 & 67.1 & 74.3\\
 & SoO Prefilling & 76.9 & 72.5 & 77.4 & 71.8 & 76.1 & 74.2 & 76.5 & 74.4 & 68.1 & 77.0\\
\bottomrule
\end{tabular}
\caption{Second-order ToM Performance (\%) for each factor of the big five personality traits of characters. For each factor (O=openness to experience, C=conscientiousness, E=extraversion, A=agreeableness, N=neuroticism), the scores on two subsets (the corresponding factor is high and low) averaged over three runs are reported.}
\label{tab:results-second-personality}
\end{table*}

\clearpage

\section{Analysis}
\label{app:analysis}
\subsection{Thought Length}
\begin{figure}[h]
    \centering
    \begin{tabular}{c}
    \includegraphics[width=5cm]{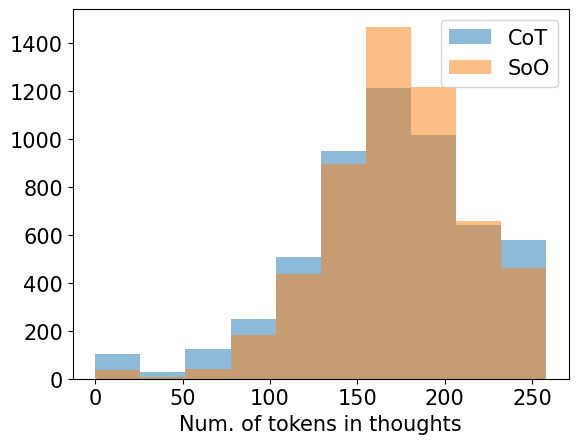} \\
    (a) ToMATO \\
    \includegraphics[width=5cm]{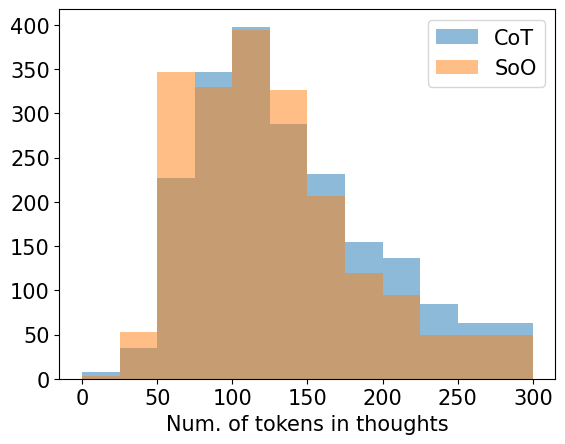} \\
    (b) ToMBench \\
    \end{tabular}
    \caption{Distributions of the number of tokens in thoughts generated with CoT and SoO.}
    \label{fig:length-dist}
\end{figure}

Table \ref{tab:stats-length} and Figure \ref{fig:length-dist} show the statistics and distributions of thought lengths, respectively.
We used the Llama-3-8B-Instruct tokenizer to count tokens.
The number of tokens in thoughts generated by SoO prefilling is generally larger than CoT prefilling on ToMATO but is smaller than CoT prefilling on ToMBench.
Meanwhile, SoO prefilling improved ToM performance on both ToMATO and ToMBench.
These findings also support that SoO prefilling does not rely solely on lengthening thoughts and scaling inference costs \cite{snell2024scaling} to improve ToM performance.

\begin{table}[h]
    \centering
    \small
    \begin{tabular}{c|cc}
    \toprule
     & ToMATO & ToMBench \\
    \midrule
    CoT Prefilling & 168.4{\scriptsize $\pm$ 51.5} & 134.9{\scriptsize $\pm$ 59.4} \\
    SoO Prefilling & 173.0{\scriptsize $\pm$ 42.6} & 125.0{\scriptsize $\pm$ 56.7} \\
    \bottomrule
    \end{tabular}
    \caption{Mean$\pm$standard deviation of the number of tokens in generated thoughts.}
    \label{tab:stats-length}
\end{table}

\subsection{Word-level Correlation Analysis}
\label{app:z-statistics}
\begin{figure}[t]
    \centering
    \begin{tabular}{cc}
    \includegraphics[width=5cm]{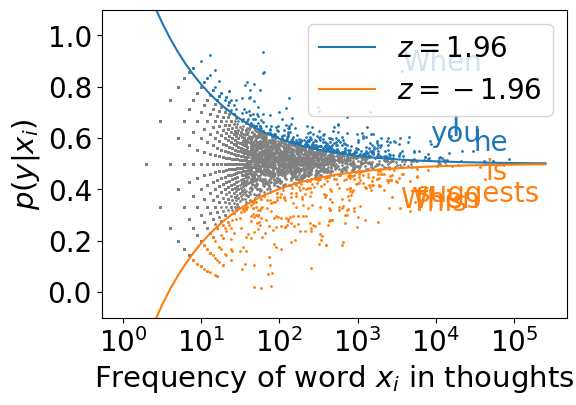} \\
    (a) ToMATO \\
    \includegraphics[width=5cm]{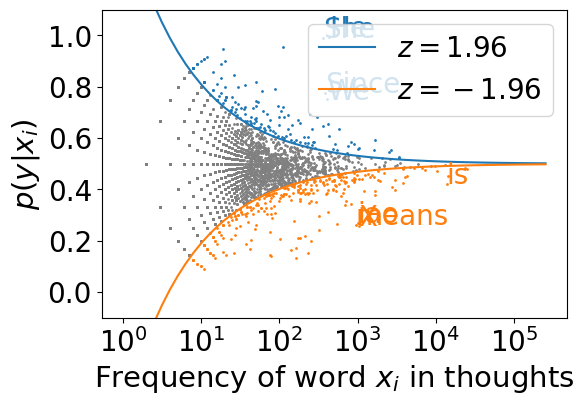} \\
    (b) ToMBench \\
    \end{tabular}
    \caption{Statistical token-level correlation analysis in thoughts generated by Llama3 7B.}
    \label{fig:z-statistics}
\end{figure}

To investigate whether the words in thoughts generated by SoO and CoT prefilling are statistically significantly different from each other, we conducted z-statistics analysis \cite{gardner-etal-2021-competency}.
Namely, we plot frequencies of word $x_i$ in thoughts and the probability $p(y|x_i)$ that word $x_i$ appears in thoughts generated by SoO.
Figure \ref{fig:z-statistics} shows the results.
Colored points represent words that appear in thoughts generated by SoO prefilling significantly higher or lower than the chance rates (1/2).
These results suggest that a substantial number of words is correlated with SoO or CoT prefilling, resulting in thoughts with different sets of words.

\subsection{Correlation Analysis of Accuracy, Faithfulness, and Thought Length}
\label{app:acc-winrate-length}

\subsubsection{Accuracy vs. Faithfulness}
\label{app:acc-winrate}
In addition to Figure \ref{fig:acc-winrate}, as shown in Figures \ref{fig:acc-winrate-mistral} and \ref{fig:acc-winrate-llama3-70b}, the SoO prefilling's win rate w.r.t. faithfulness seems to have a positive impact on the accuracy increase from CoT to SoO prefilling, except for some outliers such as the Intention (2nd, FB) subset of ToMATO.

\def\Height{3cm}
\begin{figure}[ht]
    \centering
    \begin{tabular}{c}
    \includegraphics[height=\Height]{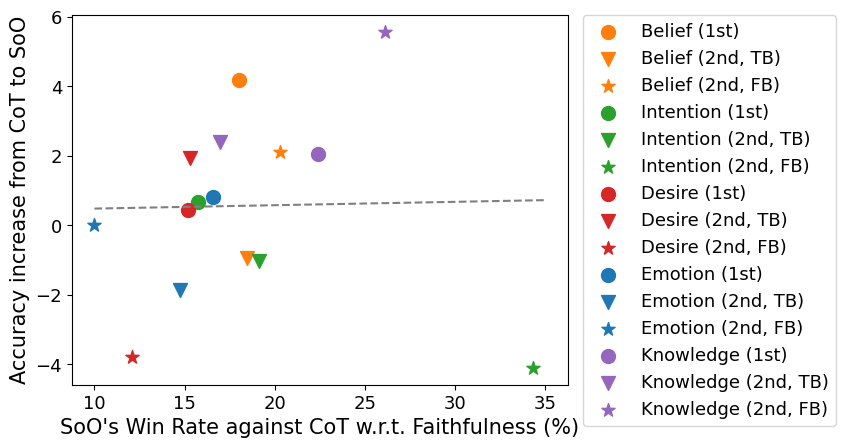} \\
    (a) ToMATO \\
    \includegraphics[height=\Height]{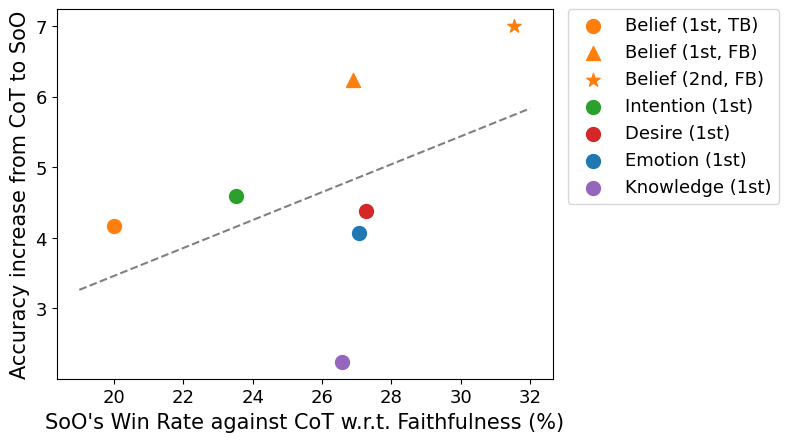} \\
    (b) ToMBench \\
    \end{tabular}  
    \caption{Correlation analysis of accuracy and faithfulness for Mistral-7B-Instruct-v0.3. The correlation between the two is positive.}
    \label{fig:acc-winrate-mistral}
\end{figure}

\begin{figure}[ht]
    \centering
    \begin{tabular}{c}
    \includegraphics[height=\Height]{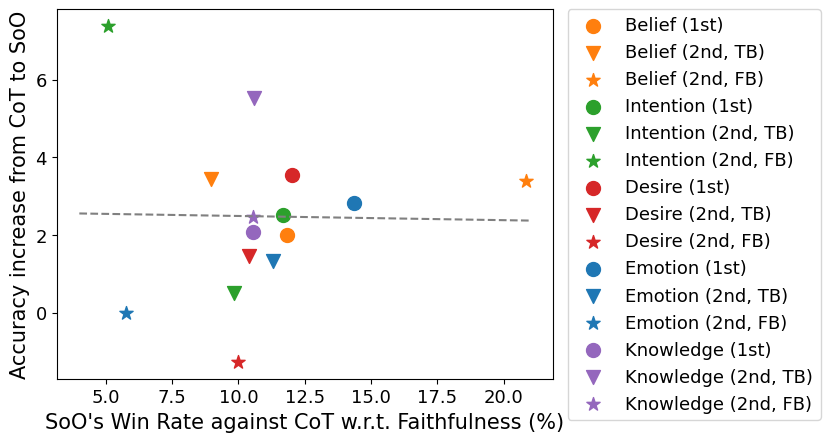} \\
    (a) ToMATO \\
    \includegraphics[height=\Height]{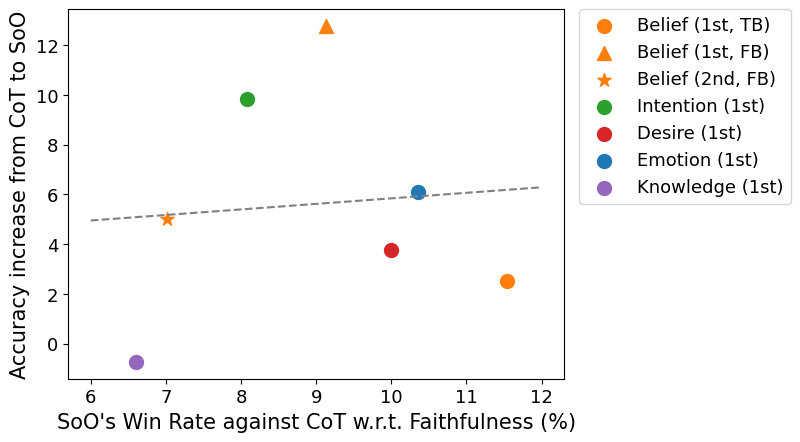} \\
    (b) ToMBench \\
    \end{tabular}  
    \caption{Correlation analysis of accuracy and faithfulness for Llama-3-70B-Instruct. The correlation between the two is positive except for some outliers such as Intention (2nd, FB).}
    \label{fig:acc-winrate-llama3-70b}
\end{figure}

\subsubsection{Accuracy vs. Thought Length}
\label{app:acc-length}
\begin{figure}[ht]
    \centering
    \begin{tabular}{c}
    \includegraphics[height=\Height]{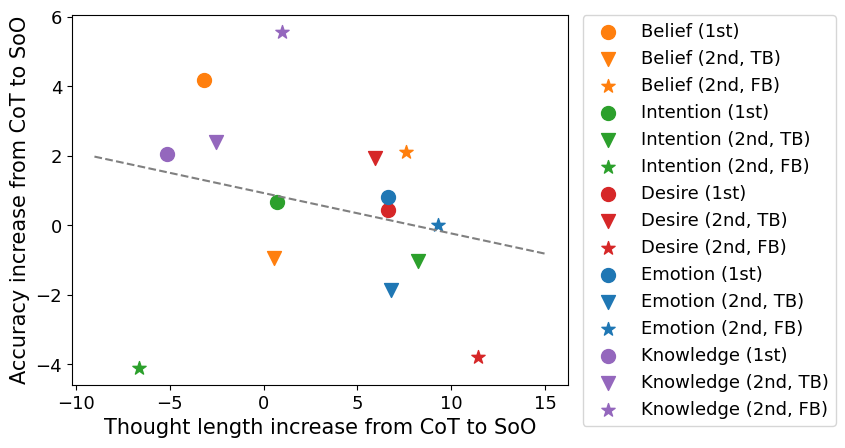} \\
    (a) ToMATO \\
    \includegraphics[height=\Height]{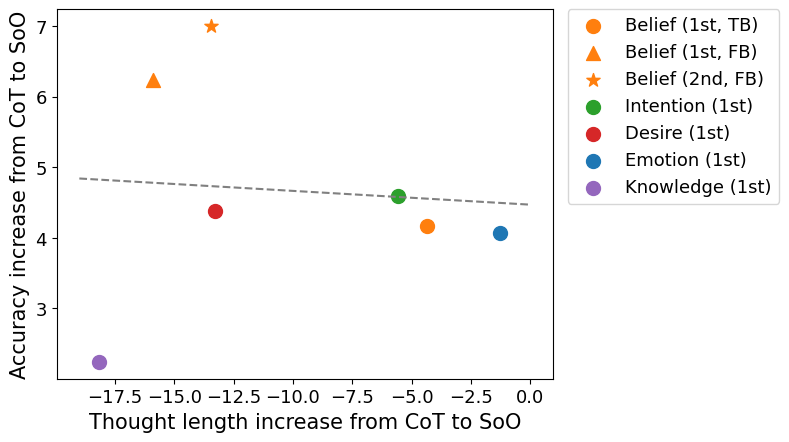} \\
    (b) ToMBench \\
    \end{tabular}  
    \caption{Correlation analysis of accuracy and length for Mistral-7B-Instruct-v0.3. The correlation between the two is not necessarily positive.}
    \label{fig:acc-length-mistral}
\end{figure}

\begin{figure}[ht]
    \centering
    \begin{tabular}{c}
    \includegraphics[height=\Height]{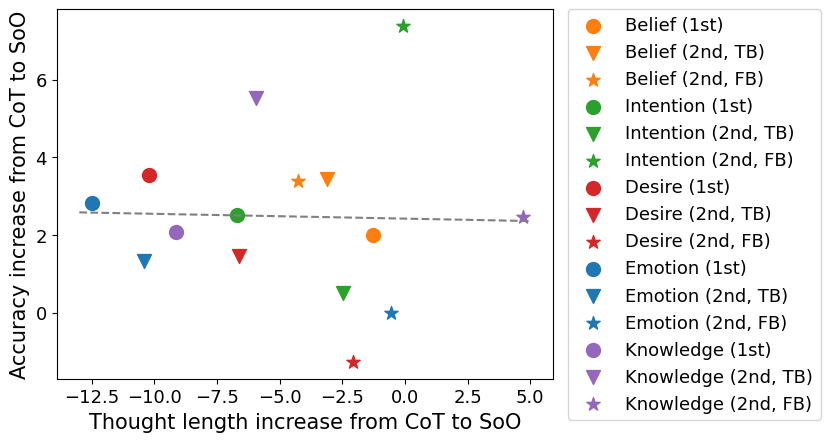} \\
    (a) ToMATO \\
    \includegraphics[height=\Height]{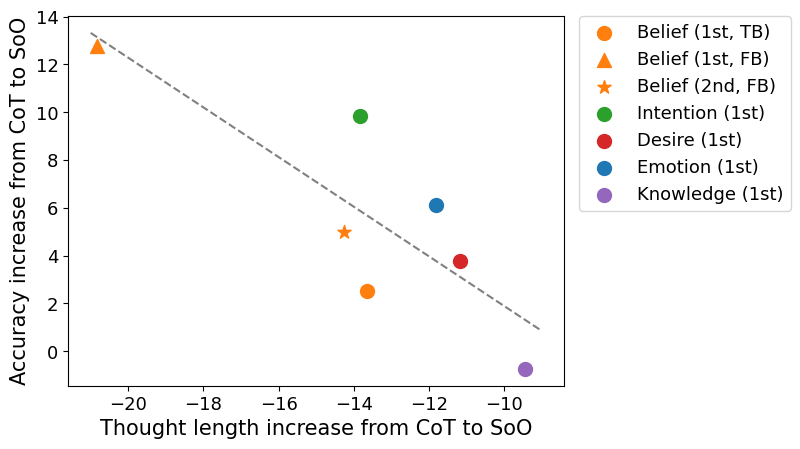} \\
    (b) ToMBench \\
    \end{tabular}  
    \caption{Correlation analysis of accuracy and length for Llama-3-70B-Instruct. The correlation between the two is not necessarily positive.}
    \label{fig:acc-length-llama3-70b}
\end{figure}

In addition to Figure \ref{fig:acc-length}, as shown in Figures \ref{fig:acc-length-mistral} and \ref{fig:acc-length-llama3-70b}, the accuracy increase is not necessarily correlated with the length increase from CoT to SoO prefilling positively.
These results also support that SoO prefilling does not rely solely on scaling the test-time compute to improve ToM.

\subsubsection{Faithfulness vs. Thought Length}
\label{app:winrate-length}

\begin{figure}[ht]
    \centering
    \begin{tabular}{c}
    \includegraphics[height=\Height]{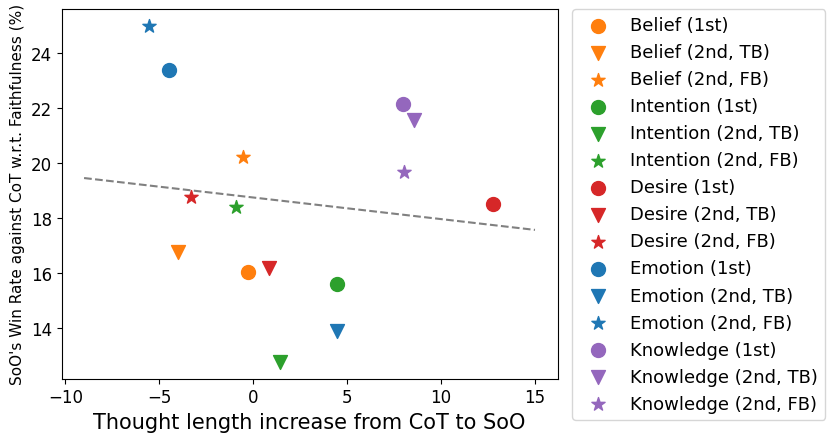} \\
    (a) ToMATO \\
    \includegraphics[height=\Height]{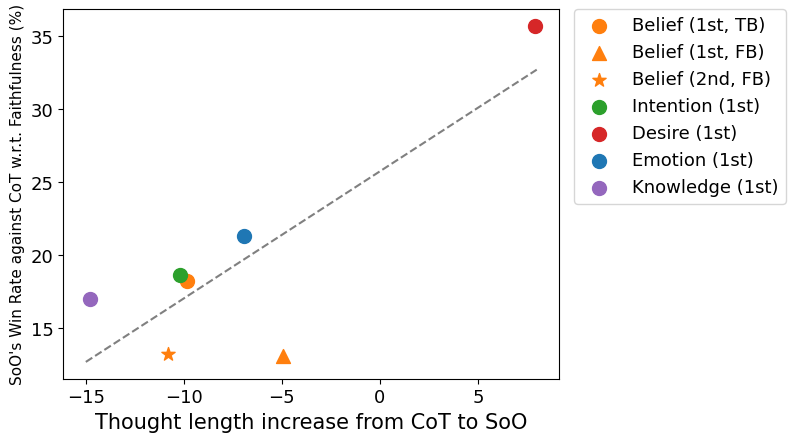} \\
    (b) ToMBench \\
    \end{tabular}    
    \caption{Correlation analysis of faithfulness and thought length for Llama-3-8B-Instruct. The correlation between the two is not necessarily positive.}
    \label{fig:winrate-length}
\end{figure}

\begin{figure}[ht]
    \centering
    \begin{tabular}{c}
    \includegraphics[height=\Height]{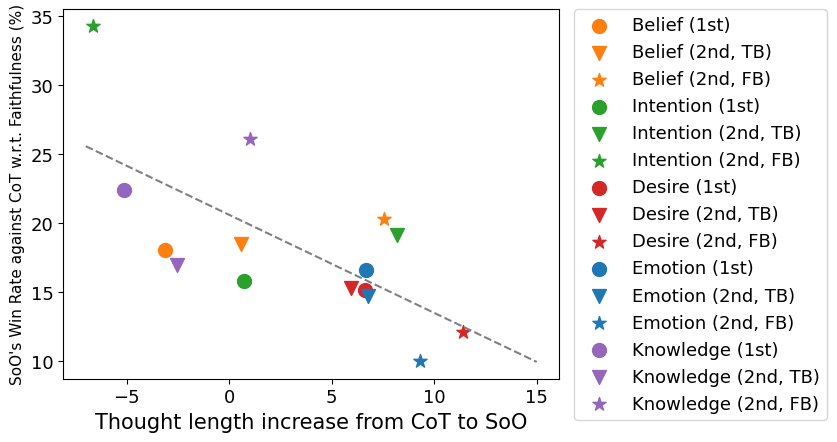} \\
    (a) ToMATO \\
    \includegraphics[width=6cm]{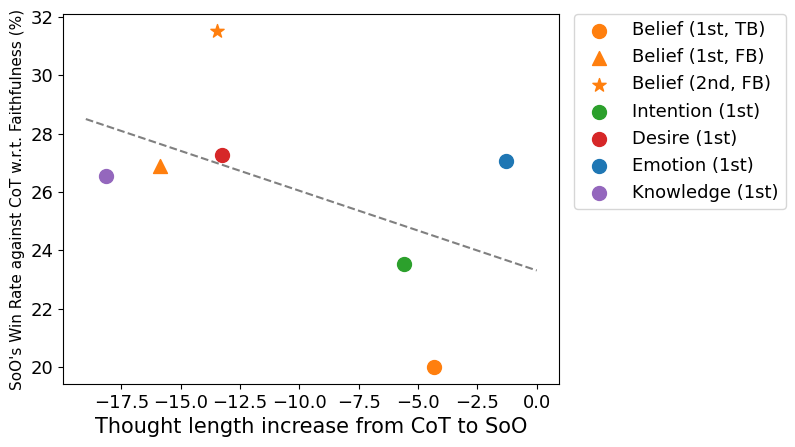} \\
    (b) ToMBench \\
    \end{tabular}  
    \caption{Correlation analysis of faithfulness and length for Mistral-7B-Instruct-v0.3. The correlation between the two is negative.}
    \label{fig:winrate-length-mistral}
\end{figure}

\begin{figure}[ht]
    \centering
    \begin{tabular}{c}
    \includegraphics[height=\Height]{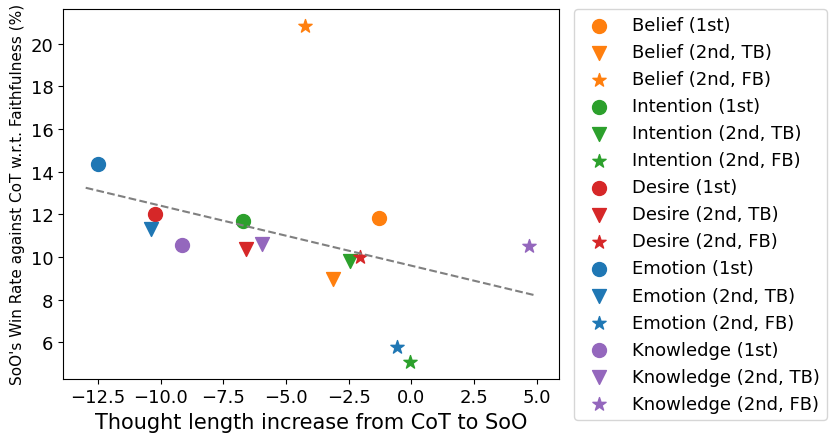} \\
    (a) ToMATO \\
    \includegraphics[width=6cm]{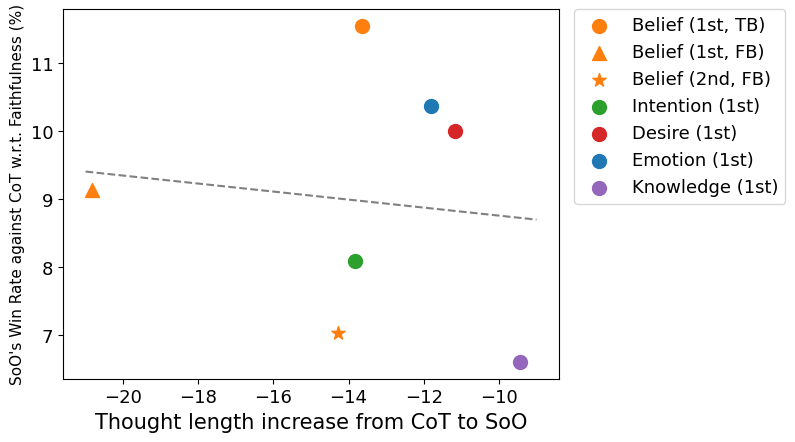} \\
    (b) ToMBench \\
    \end{tabular}  
    \caption{Correlation analysis of faithfulness and length for Llama-3-70B-Instruct. The correlation between the two is negative.}
    \label{fig:winrate-length-llama3-70b}
\end{figure}

As shown in Figures \ref{fig:winrate-length}, \ref{fig:winrate-length-mistral}, and \ref{fig:winrate-length-llama3-70b}, we showed that the thought lengths are not positively correlated with the win rates in most cases.
This suggests that LLM-as-a-judge did not suffer from verbosity bias \cite{zheng2023judging} in our analyses, i.e., the evaluator did not favor longer outputs regardless of the contents.
These results also imply that the shorter a thought, the more faithful it is in ToM tasks.

\clearpage

\subsection{Pairwise Comparison w.r.t. Faithfulness}
\label{app:pairwise}
The win, tie, lose rates w.r.t. faithfulness on ToMATO and ToMBench are shown in Table \ref{tab:winrate-all}.
The detailed results for Llama3 8B are given in Tables \ref{tab:all-winrate-tomato} and \ref{tab:all-winrate-tombench}.
These results show that SoO prefilling's win rate against CoT prefilling was larger than the lose rate in most cases.
In some cases, such as Llama3 70B on ToMBench, and Llama3 8B on Intention (2nd, TB) of ToMATO, the lose rates were larger than the win rates.
Therefore, factors other than faithfulness may also affect the superior performance of SoO prefilling.
Identifying other factors for improving ToM is future work.

\begin{table}[h]
\centering
\small
\begin{tabular}{c|ccc}
\toprule
Model & Win & Tie & Lose\\
\midrule
\multicolumn{4}{c}{ToMATO}\\
\midrule
Mistral 7B & 17.53 & 66.43 & 16.04\\
Llama3 8B & 18.22 & 67.51 & 14.27 \\
Llama3 70B & 11.28 & 79.49 & 9.23 \\
\midrule
\multicolumn{4}{c}{ToMBench}\\
\midrule
Mistral 7B & 25.62 & 53.94 & 20.45\\
Llama3 8B & 19.20 & 66.77 & 14.04 \\
Llama3 70B & 9.52 & 78.03 & 12.45\\
\bottomrule
\end{tabular}
\caption{SoO prefilling's win/tie/lose rate (\%) against CoT prefilling w.r.t. faithfulness on ToMATO and ToMBench.}
\label{tab:winrate-all}
\end{table}

\begin{table}[h]
    \centering
    \small
    \begin{tabular}{c|ccc}
    \toprule
    Subset & Win & Tie & Lose \\
    \midrule
    Belief (1st) & 16.04 & 66.21 & 17.75\\
    Belief (2nd, TB) & 16.78 & 66.44 & 16.78\\
    Belief (2nd, FB) & 20.24 & 63.09 & 16.67\\
    Intention (1st) & 15.60 & 72.78 & 11.62\\
    Intention (2nd, TB) & 12.76 & 70.40 & 16.84\\
    Intention (2nd, FB) & 18.42 & 71.05 & 10.53\\
    Desire (1st) & 18.51 & 68.68 & 12.81\\
    Desire (2nd, TB) & 16.18 & 75.00 & 8.82\\
    Desire (2nd, FB) & 18.75 & 56.25 & 25.00\\
    Emotion (1st) & 23.40 & 63.77 & 12.83\\
    Emotion (2nd, TB) & 13.87 & 73.99 & 12.14\\
    Emotion (2nd, FB) & 25.00 & 55.36 & 19.64\\
    Knowledge (1st) & 22.15 & 62.28 & 15.57\\
    Knowledge (2nd, TB) & 21.57 & 64.70 & 13.73\\
    Knowledge (2nd, FB) & 19.67 & 67.22 & 13.11\\
    \midrule
    ALL & 18.22 & 67.51 & 14.27 \\
    \bottomrule
    \end{tabular}
    \caption{SoO prefilling's win/tie/lose rate (\%) against CoT prefilling w.r.t. faithfulness for each subset of ToMATO.}
    \label{tab:all-winrate-tomato}
\end{table}

\begin{table}[h]
    \centering
    \small
    \begin{tabular}{c|ccc}
    \toprule
    Subset & Win & Tie & Lose \\
    \midrule
    Belief (1st, TB) & 18.18 & 65.15 & 16.67\\
    Belief (1st, FB) & 13.08 & 72.30 & 14.62\\
    Belief (2nd, FB) & 13.19 & 75.82 & 10.99\\
    Intention (1st) & 18.60 & 66.87 & 14.53\\
    Desire (1st) & 35.71 & 51.43 & 12.86\\
    Emotion (1st) & 21.28 & 65.10 & 13.62\\
    Knowledge (1st) & 16.95 & 64.41 & 18.64\\
    \midrule
    ALL & 19.20 & 66.77 & 14.04 \\
    \bottomrule
    \end{tabular}
    \caption{SoO prefilling's win/tie/lose rate (\%) against CoT prefilling w.r.t. faithfulness for each subset of ToMBench.}
    \label{tab:all-winrate-tombench}
\end{table}

\newpage

\section{Examples}
\label{app:example}
Examples of input and output from ToMATO and ToMBench are presented in Tables \ref{tab:example-tomato} and \ref{tab:example-tombench}, respectively.

\begin{table*}[h]
\small
\centering
\begin{tabular}{p{15cm}}
\toprule
Input\\
\midrule
\# Conversation\\
Liam Johnson: "Hi!"\\
Ethan Johnson: "Hi, how are you?"\\
Liam Johnson: "Ethan, I've noticed you've been a bit distant lately, is everything okay?"\\
Ethan Johnson: "Yeah, everything's fine, just been really busy with the restaurant and some personal stuff, you know how it is."\\
Liam Johnson: "Come on, Ethan, we've been friends for years, I can tell when something's bothering you. What's really going on?"\\
Ethan Johnson: "Alright, alright... to be honest, I've been having some trouble making ends meet, the restaurant's been slow and I'm falling behind on some bills, but I'll figure it out, I always do."\\
Liam Johnson: "Ethan, you're not going to 'figure it out' on your own, you need a plan and some guidance. Let me take a look at your finances and we can go over some options together, okay?"\\
Ethan Johnson: "Man, I appreciate that, Liam, but I don't want to be a burden, I can handle it, I just need to get my head out of my... okay, fine, yeah, maybe a fresh set of eyes would help, thanks."\\
Liam Johnson: "Alright, let's get down to business then. I'll take a look at your books and we'll get a plan in place. But next time, don't be so stubborn and come to me sooner, got it?"\\
Ethan Johnson: "Ha! Yeah, yeah, got it, I'll try to work on that, thanks, Liam. And thanks again, man, I owe you one... or a few thousand, it seems."\\
Liam Johnson: "Yeah, yeah, just pay me back in free meals at the restaurant, and we'll call it square. But seriously, let's take a close look at your expenses and see where we can cut back, okay?"\\
Ethan Johnson: "Aww, man, you're a lifesaver, Liam. Yeah, let's dig in and see where I can tighten the belt, I'm sure there's some fat to trim. And don't worry, you'll be eating like a king at my place for a long time to come."\\
Liam Johnson: "Alright, let's get to work then. I'll start going over the numbers and we can discuss some options. And don't make promises you can't keep, just focus on getting back on your feet, okay?"\\
Ethan Johnson: "Yeah, yeah, got it, thanks for keeping me grounded, Liam. I'll let you do your magic with the numbers, and I'll focus on getting the restaurant back on track. You're right, one step at a time."\\
Liam Johnson: "That's the spirit, Ethan. Just remember, I'm not a miracle worker, you need to put in the effort too. Let's schedule a follow-up meeting for next week and we'll reassess where we're at, sound good?"\\
Ethan Johnson: "Absolutely, sounds like a plan, Liam. I'll make sure to have all my ducks in a row by then. And thanks again, man, just knowing you've got my back makes me feel a hell of a lot better."\\
\# Question\\
When Ethan says "Yeah, everything's fine, just been really busy with the restaurant and some personal stuff, you know how it is.", how does he think that Liam feels?\\
\# Options\\
{[A]} He thinks that he feels \textcolor{blue}{concerned} and genuinely wants to help, but also might be a bit worried about getting involved in his problems\\
{[B]} He thinks that he feels empathetic and is pushing him to open up because he cares about him, but also might be a bit frustrated that he's not being honest with him\\
{[C]} He thinks that he feels a sense of optimism and encouragement, like he's confident that they can get through this together\\
{[D]} He thinks that he feels a sense of relief that he's finally opening up and accepting his help, but also a hint of amusement at his stubbornness\\
\midrule
Chain-of-Thought prefilling\\
\midrule
Let's think step by step. When Ethan says ``Yeah, everything's fine, just been really busy with the restaurant and some personal stuff, you know how it is.'', he is trying to brush off Liam's concern and make it seem like everything is okay. This suggests that he thinks Liam is \textcolor{blue}{concerned} about his situation and is trying to downplay it.\\
Option {[B]} seems the most probable answer. Ethan thinks that Liam feels \textcolor{red}{empathetic} and is pushing him to open up because he cares about him. This is evident when Liam says ``Come on, Ethan, we've been friends for years, I can tell when something's bothering you.'' Liam is genuinely concerned about Ethan's well-being and is trying to get to the bottom of the issue.\\
\midrule
Shoes-of-Others prefilling\\\midrule
Let's put ourselves in Ethan's shoes. When he says ``Yeah, everything's fine, just been really busy with the restaurant and some personal stuff, you know how it is.'', he's trying to brush off Liam's concern and maintain a facade of normalcy. He's using a generic excuse to deflect Liam's probing questions and avoid discussing his actual problems.\\
Given this context, it's likely that Ethan thinks Liam feels \textcolor{blue}{concerned} and genuinely wants to help, but also might be a bit worried about getting involved in his problems. This is why Ethan's response is a bit evasive and tries to downplay the issue. He's not entirely wrong, as Liam does express concern and a willingness to help, but Ethan is also aware that Liam might be hesitant to get too deeply involved in his personal life.\\
So, the most probable answer is:\\
{[A]} He thinks that he feels \textcolor{blue}{concerned} and genuinely wants to help, but also might be a bit worried about getting involved in his problems\\
\bottomrule
\end{tabular}
\caption{An example of the input from ToMATO and the outputs of Llama-3-8B-Instruct with Chain-of-Thought and Shoes-of-Others prefilling. CoT prefilling generated unfaithful thoughts, i.e., it generated ``he (Ethan) thinks that Liam is \textcolor{blue}{concerned},'' but predicted the wrong answer, [B] Ethan thinks that Liam feels \textcolor{red}{empathetic}.}
\label{tab:example-tomato}
\end{table*}

\begin{table*}[h]
\small
\centering
\begin{tabular}{p{15cm}}
\toprule
Input\\
\midrule
\# Context\\
Sara is a designer, she always hopes to design a unique LOGO for the company's new project. However, her teammate is on sick leave recently, the task of making the promotional video falls on her, she decides to fully complete the production of the promotional video.\\
\# Question\\
What does Sara do after she completes the production of the promotional video?\\
\# Options\\
{[A]} Sara continues to complete the production of the project logo.\\
{[B]} Sara continues to produce more promotional videos.\\
{[C]} Sara takes a break for a while and then carries out new tasks.\\
{[D]} Sara resigns from her position as a designer.\\
\midrule
Chain-of-Thought prefilling\\
\midrule
Let's think step by step. Yang Hui whispers something to Xiao Lin, and only the two of them know what it is. Then, they look at each other and laugh. This suggests that the reason for their laughter is not something that can be observed by others, like the planning of the volunteer activity.\\
After the ``whisper'', Yang Hui invites Xiao Wang to participate in the volunteer activity, and Xiao Wang remembers their smiles and eye contact. This implies that the content of the ``whisper'' is related to Xiao Wang.\\
Given these clues, the most probable answer is [D] Yang Hui and Xiao Lin discuss the interesting things about Xiao Wang, and they smile because of this. The fact that Yang Hui invites Xiao Wang to participate in the volunteer activity after the ``whisper'' suggests that the topic of the ``whisper'' is related to Xiao Wang's capabilities or qualifications for the activity.\\
\midrule
Shoes-of-Others prefilling\\
\midrule
Let's put ourselves in Yang Hui's shoes. Since Yang Hui whispers something to Xiao Lin, it's likely that the information shared is not publicly known, and Xiao Lin is not supposed to share it with anyone else. The fact that they look at each other and laugh suggests that the information is amusing or surprising, and they're sharing a private joke.\\
The fact that Yang Hui then leads Xiao Wang to participate in the volunteer activity, and Xiao Wang remembers their smiles and eye contact, implies that the information shared is relevant to Xiao Wang's involvement in the activity.\\
Considering these clues, I believe the most probable answer is:\\
{[C]} Yang Hui and Xiao Lin both think that Xiao Wang is capable of organizing this activity, so they laugh.\\
The whisper likely contains some information about Xiao Wang's abilities or strengths that make Yang Hui and Xiao Lin excited about involving him in the volunteer activity.\\
\bottomrule
\end{tabular}
\caption{An example of the input from ToMBench and the outputs of Llama-3-8B-Instruct with Chain-of-Thought and Shoes-of-Others prefilling.}
\label{tab:example-tombench}
\end{table*}

\end{document}